\documentclass[draft]{article} 
\usepackage{lambek-fs}
\usepackage{ushort} 

\newcommand\fv{f\!dv}
\newcommand\dland{\mathbin{\ushortw{{\land}}}}

\newcommand\precm{\mathrel{\prec}^{-}}
\newcommand\precp{\mathrel{\prec}^{+}}

\newcommand\ante{\nwarrow}

\newcommand\formcmg{\mathcal{F}}
\newcommand\mgentries{\mathcal{L}}

\newcommand\monte{\mathbf{e}}
\newcommand\montt{\mathbf{t}}
\newcommand\montev{\mathbf{v}}

\newcommand\Lex{\mathrm{Lex}}

\newcommand\lto{\mathbin{\backslash}}
\newcommand\lfrom{\mathbin{/}}

\newcommand\llts{\mathbin{\otimes}}

\newcommand\seq\vdash 
\newcommand\fl{\rightarrow} 

\newcommand\hdrl\Rightarrow
\newcommand\hdr\Leftarrow

\newcommand\ltohdr{\mathbin{_{\hdrl}\!\backslash}}

\newcommand\lfromhdr{\mathbin{/\!_{\hdr}}} 

\newcommand\base{\mathsf{b}}
\newcommand\demand{\mathsf{^{=}b}}
\newcommand\head{\mathsf{b^{\hdr}}}
\newcommand\move{\mathsf{{-}m}}
\newcommand\place{\mathsf{{+}m}}

\newcommand\simpl\varsigma 

\usepackage{prooftree}

\newcommand\pheat{\mathrm{eat}}
\newcommand\children{\mathrm{children}}
\newcommand\les{\mathrm{the}}
\newcommand\pizza{\mathrm{pizza}} 
\newcommand\ate{\mathrm{ate}}
\newcommand\un{\mathrm{a}}

\newcommand\past{\mathtt{P}}
\newcommand\agent{\mathtt{Ag}}
\newcommand\patient{\mathtt{Pa}}

\newcommand\child{\mathtt{child}}
\newcommand\pizz{\mathtt{piz}} 
\newcommand\eat{\mathtt{eat}}

\newcommand\infl{\mathit{infl}}
\newcommand\modif{\mathit{modif}}
\newcommand\comp{\mathit{comp}}

\newcommand\rien{\varepsilon}

\newcommand{\labitem}[6]{
\begin{array}{l} 
#1\\ 
#2\\ 
\end{array} 
\vdash \left\{
\begin{array}{l}
#3\\ 
{\sf #4}\\ 
{#5}\\ 
\hspace*{-1ex}\begin{array}{l}#6\end{array}\\  
\end{array}\right.}

\newcommand\derivelem[4] 
{\bigskip\noindent\begin{prooftree} 
#1\quad #2
\justifies 
#3
\using{_{#4}}
\end{prooftree}\bigskip}

\newcommand\aaa{\labitem
{}{}{(\rien\,|\,\pheat\,|\,\rien)}
{V\lfrom d}
{\monte \rightarrow \monte \rightarrow \montev \rightarrow \montt}
{\lambda x \lambda y \lambda e . \eat(e,x,y)}
} 

\newcommand\bbb{\labitem{u{:}\sf d}{\underline u{:} \monte}{(\rien\,|\,u\,|\,\rien)}
{d}{\monte}{\underline u}
}

\newcommand\ccc{\labitem{u{:}\sf d}{\underline u{:} \monte}{(\rien\,|\,\pheat\,|\,u)}
{V}{\monte \rightarrow \montev \rightarrow \montt}
{\lambda y \lambda e .\\ \eat(e,\underline u,y)}
}

\newcommand\ddd{\labitem{}{}{(\rien\,|\,\rien\,|\,\rien)}
{(k\backslash (d \backslash v))\lfromhdr V}
{( \monte {\rightarrow} \montev {\rightarrow} \montt) {\rightarrow} \monte {\rightarrow} \monte {\rightarrow} \montev {\rightarrow} \montt}
{ \lambda R \lambda x_2 \lambda y \lambda e .\\ R (y,e) \dland \patient(e,x_2)}
}

\newcommand\eee{\labitem{u{:}\sf d}{\underline u{:} \monte}{(\rien\,|\,\pheat\,|\,u)}
{ k\backslash (d \backslash v)}
{\monte \rightarrow \monte \rightarrow \montev \rightarrow \montt} 
{\lambda x_2 \lambda y \lambda e . \eat(e,\underline u, y)\\ \dland \patient(e,x_2)}
} 

\newcommand\fff{\labitem{v{:}\sf k}{\underline v{:} \monte}{(\rien\,|\,v\,|\,\rien)}
{k}{\monte}{\underline v}
}

\newcommand\gggg{\labitem{v{:}\sf k,u{:}\sf d}{\underline v{:} \monte, \underline u{:} \monte}{(v\,|\,\pheat\,|\,u)}
{d \backslash v}
{\monte \rightarrow \montev \rightarrow \montt}
{\lambda y \lambda e .\\ \eat(e,\underline u,y)\\ {\dland} \patient(e,\underline v) }
}

\newcommand\hhh{\labitem{}{}{(\rien\,|\,\un\,|\,\rien)}
{(k \otimes d)\lfrom n}{(\monte \rightarrow \montt) \rightarrow \monte}
{\lambda Q. \mu \gamma . [p \,|\, Q(p) \land \gamma(p)]}
}

\newcommand\iii{\labitem{}{}{(\rien\,|\,\pizza\,|\,\rien)}
{n}
{\monte\rightarrow \montt}
{\lambda z. \pizz(z)}
}

\newcommand\kkk{\labitem{}{}{(\rien\,|\,\un\,|\,\pizza)}
{k \otimes d}
{\monte}
{\mu \gamma .\\{}[p \,|\, \pizz(p){\land}\gamma(p)]}
}

\newcommand\llll{\labitem{}{}{(\un\,\pizza\,|\,\pheat\,|\,\rien)}
{d \backslash v}
{\monte \rightarrow \montev \rightarrow \montt}
{\lambda y \lambda e .\\ \eat(e,\mu \gamma . [p \,|\, \pizz(p){\land} \gamma(p)],y)\\ {\dland} \patient(e,p)}
}

\newcommand\mmm{\labitem{w{:}\sf d}{\underline w{:} \monte}{(\rien\,|\,w\,|\,\rien)}
{d}{\monte}{\underline w}
}

\newcommand\nnn{\labitem{w{:}\sf d}{\underline w{:} \monte}{(w\ \un\,\pizza\,|\,\pheat\,|\,\rien)}{v}{\montev \rightarrow \montt}{ \lambda e .\\ \eat(e,\mu \gamma .[p \,|\, \pizz(p) {\land} \gamma(p)],\\ \underline w) {\dland} \patient(e,p)}}

\newcommand\ooo{\labitem
{}{}{(\rien\,|\,\infl\,|\,\rien)}
{(k \lto t )\lfromhdr v}
{(\montev {\rightarrow} \montt) {\rightarrow} \monte {\rightarrow} \montev {\rightarrow} \montt}
{\lambda Q \lambda y_2 \lambda e .\\ Q(e) {\dland} \past(e)\\ {\land} agent(e, y_2)} 
}

\newcommand\ppp{\labitem
{w{:}\sf d}{\underline w{:} \monte}{(\rien\,|\, \ate\,|\, w\ \un\ \pizza)}
{k \lto t}
{\monte \rightarrow \montev \rightarrow \montt}
{\lambda y_2 \lambda e .\\ \eat(e,\mu \gamma . [p \,|\, \pizz(p) {\land} \gamma(p)],\underline w)\\ {\dland} \patient(e,p) {\dland} \past(e)\\ {\dland} \agent(e, y_2)}
}
\newcommand\qqq{\labitem{y{:}k}{\underline y{:}\monte}{(\rien\,|\,y\,|\,\rien)}{k}{\monte}{\underline y}}

\newcommand\rrr{\labitem
{w{:}\sf d,y{:}\sf k}{\underline w{:} \monte, \underline y{:} \monte}{(y\,|\,\ate\,|\,w\ \un\,\pizza)} 
{t}{\montev \rightarrow \montt}
{\lambda e .\\ \eat(e,\mu \gamma .\\ \ [p \,|\, \pizz(p) {\land} \gamma(p)],\\ \underline w)\\ {\dland}  \patient(e,p){\dland} \past(e)\\ {\dland} \agent(e, \underline y)} 
}

\newcommand\sss{\labitem
{}{}{(\rien\,|\,\les\,|\,\rien)}
{(k \otimes d) \lfrom n}
{(\monte \rightarrow \montt) \rightarrow \monte}
{\lambda Q \mu \delta . [\,|\,[d\,|\, Q(d)] {\Rightarrow} [\delta(d)]]}
}

\newcommand\ttt{\labitem{}{}{(\rien\,|\,\children\,|\,\rien)}
{n}{\monte\rightarrow \montt}
{\lambda z. \child(z)}
}

\newcommand\uuu{\labitem{}{}{(\rien\,|\,\les\,|\,\children)}
{k \otimes d} 
{\monte}
{\mu \delta . [\,|\,[d\,|\, \child(d)]\\ \quad {\Rightarrow} [\,|\,\delta(d)]]} 
}

\newcommand\vvv{\labitem
{}{}{(\les\,\children\,|\,\ate\,|\,\un\,\pizza)}
{t} 
{\montev \rightarrow \montt}
{\lambda e. \eat(e,\mu \gamma . [p \,|\, \pizz(p) \land \gamma(p)], \\ \hfill \mu \delta . [\,|\,[d\,|\, \child(d)] {\Rightarrow} [\,|\,\delta(d)]])\\ {\dland} \patient(e,p) {\dland} \past(e) {\dland} \agent(e, d)}
}

\newcommand\xxx{\labitem 
{}{}{(\rien\,|\,\rien\,|\,\rien)} 
{C \lfrom  t} 
{(\montev \rightarrow \montt) \rightarrow \montt}
{\lambda Q . [e\,|\, (Q(e))]}
}

\newcommand\yyy{\labitem {}{}
{(\rien\,|\,\rien\,|\,\les\,\children\,\ate\,\un\,\pizza)}
{C}
{\montt}
{[e\,|\,\eat(e,\mu \gamma . [p \,|\, \pizz(p) \land \gamma(p)], \\ \hfill \mu \delta . [\,|\,[d\,|\, \child(d)] {\Rightarrow} [\,|\,\delta(d)]])\\ {\dland} \patient(e,p) {\dland} \past(e) {\dland} \agent(e, d)]}
}

\title{Categorial minimalist grammars:\\ from generative syntax to logical forms}
\author{Maxime Amblard, Alain Lecomte, Christian Retor\'e}
\institution{Nancy Universit\'e, Universit\'e Paris 8, Universit\'e de Bordeaux}
\begin{document}
\maketitle


\section{Convergence at first sight,\\ and differences in a second time}

From the early days of the minimalist program of \cite{Cho93}, a convergence with categorial grammars was noticed by \cite{BE95}. 
The striking similarity lies in the merge operation, which looks like the application rule of the AB grammars. 
In both cases, word order is a consequence of the consumption rules
but in categorial grammars being the head of a compound expression also 
derives from the categories while in a minimalist setting it can be defined independently from the resource consumption. 
The most striking difference is  the absence of movement in categorial grammar: how could this notion be captured? If it can,  how could minimality conditions like \emph{shortest move} be formulated? 
In the other direction the main difference is the  atomicity of minimalist features: \emph{merge} is not recursive, there cannot be a demand of a category which itself demand another category. 

Hence, after quite an optimistic wish of convergence, the task seems so tough
that one may wonder why we are willing to do so? There are at least two reasons: 
\begin{itemize} 
\item 
Of course, the main reason is that categorial grammar easily yield the logical form (or semantic representation) following some of Montague's idea. 
\item A secondary reason is the formalisation of converging learning algorithms from positive structured examples, which can be defined for categorial grammars, but we shall not speak about it, 
although one of us did something in this direction \cite{BR01lll}. 
\end{itemize} 

Because the categorial tradition is anchored in a formal and logical apparatus,  the convergence between generative grammar and 
categorial grammars  benefited from the development of resource logic and 
from the formalisation of the minimalist program  by \cite{Sta96}  into tree grammars known as minimalist grammars  
as a special issue of Language and Computation shows 
(\cite{RS03}). 

The kind of categorial grammar we are using relies on the outcomes of linear logic \citep{Gir87},
which extends the Lambek calculus \citep{Lam58} which itself is a logical completion of the AB-grammars \citep{BH53}Êwhen the fraction rules are viewed as \emph{modus ponens}.
Lambek calculus is quite a restricted logic: it is an intuitionistic logic, where every hypothesis is used once and exactly once, 
and where they cannot permute. Our representation 
uses a mixed system with both commutative and non commutative connectives \citep{Gro96,Ret04petri,AR07cie}. 
Still it is a linear subsystem of intutionistic logic, and this enables to derive semantic interpretation from the syntactic analysis. 

There are several ways to represent rather faithfully Edward Stabler's minimalist grammars in categorial grammars and to endow them with an integrated treatment of compositional semantics. 
Some of them take place in the multimodal framework, like  \citep{Cor98, Ver99rlmg}, but here we focus on the one we introduced in the framework of linear logic 
\cite{LR99fg,LR01acl} that we call categorial minimalist grammars. 
We developed them with others (in particular \cite{Anoun07phd}), but 
in this presentation, we rather follow the one thoroughly presented in the PhD thesis of the first author,\cite{Amblard07phd}. 

We first recall some basic notions on minimalist grammars and on categorial grammars. 
Next we shortly introduce partially commutative linear logic, 
and our representation of minimalist grammars within this categorial system,
the so-called categorial minimalist grammars. 
Next we briefly present $\lambda\mu$-DRT (Discourse Representation Theory) an extension of $\lambda$-DRT (compositional DRT)
in the framework of $\lambda\mu$ calculus: it avoids type raising and derives different readings from a single semantic representation,
in a setting which follows discourse structure. 
We run a complete example which illustrates the various structures and rules that are needed to derive a semantic representation from the  categorial
view of a transformational syntactic analysis. 

\section{Minimalist grammars}

There are actually several variants of minimalist grammars, due to Edward Stabler, whose guidelines are similar. 
We focus on the initial formulation of \cite{Sta96}, an important alternative being \cite{Sta99} where operations that one may find unpleasant in a categorial setting like \emph{head-movement}, are simulated by sequences of elementary operations.

As usual in a lexicalised grammar, the lexicon maps each lexical item from 
$\Phi=\{Peter, sleeps, loves,...\}$ 
 to a formal encoding of its syntactic behaviour.  
 Here syntactic behaviour is depicted by a sequence of features (category, demand of a category, movement trigger, movement target, see figure \ref{mgentries}) followed by the corresponding phonological form. This sequence controls  the generating process,  which makes use of independent generating functions,  only depending on the features of the heads.

\begin{figure} 
\begin{center} 
$$
\begin{array}{rclp{0.4\textwidth}} 
\multicolumn{4}{c}{\mbox{Minimalist features}}\\ 
\base&=&\sf \{C,v, V, d, t, ...\} & base syntactic categories\\ 
\demand&=&\{{=}x | x\in \base\} & demand of a base category, selectors\\
\head&=&\{x{\hdr} | x\in \base\} & demand of a base category, with specific placing of the head, i.e. head movement selectors\\
\move&=&\sf \{{-}k,{-}wh,...\} & movement triggers, licensees \\ 
\place&=&\{{+}x | {-}x\in\move\} & movement target, licensors\\ 
\end{array}  
$$\\Ê
\end{center} 
\caption{Features and lexical entries in minimalist grammars} 
\label{mgentries} 
\end{figure}

\begin{defn}{Minimalist lexical entries}
\label{mglexicalentries}
The sequence of features preceding the phonological form matches  the following regular expression, whose ingredients are depicted in figure \ref{mgentries}: 
$$Lex(word)\in\mgentries = (\demand\cup\head (\demand\cup\head\cup \place)^*)^* \base (\move)^*$$
\end{defn}

Such grammars operate on binary trees, the leaves of which are labelled with sequences of features, and the nodes of which are labelled by either "$<$" or "$>$". 
The \emph{head} of a tree is defined recursively as follows: the head of a tree reduced to a leaf is itself.  The head of $T_1 < T_2$ and of $T_2 > T_1$ is the head of $T_1$. 
Given a leaf $\ell$ of a tree $T$, there always exists a  biggest subtree $T'$ of $T$ such
that $l$ is the head of $T'$: $T'$ is said to be the maximal projection of $\ell$.

The \emph{initial rule} says that a tree reduced to a leaf labelled by the lexical entry is itself a derivation tree. 

The \emph{merge}  rule, is defined for two trees $T_1$ and $T_2$, one having a head $\tau_1 = {=}x \tau'_1 \phi_1$ (the demand) and the other having a head $\tau_2 = x \tau'_2 \phi_2$, where $\tau'_1,\tau'_2$ are sequences of features and $\phi_1,\phi_2$ are sequences of phonological features. 
Let us call $T'_1$ and $T'_2$ the trees obtained from $T_1$ and $T_2$ by suppressing from their respective heads the first features, that are respectively $x$ and ${=}x$. 
If the demand comes from a lexical item, then the result of the merge operation is $T'_1<T'_2$. 
Otherwise, the result is $T'_2>T'_1$. In any case the head of the result comes from $T_1$ and is $\tau'_1 \phi_1$.

The \emph{head-movement with right adjunction} is actually a special merge and not a kind of movement.  It is defined for two trees $T_1$ and $T_2$, one having a head $\tau_1 = x{\hdr} \tau'_1 \phi_1$ (the demand) and the other having a head $\tau_2 = x \tau'_2 \phi_2$, where $\tau'_1,\tau'_2$ are sequences of features and $\phi_1,\phi_2$ are sequences of phonological features. 
Let us call $T'_1$ the tree obtained from $T_1$ by turning its head into $\tau_1 = \tau'_1 \phi_1 \phi_2$ i.e. the demand $x{\hdr}$ has been suppressed 
and the phonological features are the ones of $T_1$'s head followed by the one of the head of $T_2$. 
Let us call  $T'_2$ the tree obtained from $T_2$ by turning its head into $\tau'_2$ i.e. by suppressing both the $x$ that has been consumed during the merge and the phonological features of the head that have been moved into $T'_1$. 
If the demand comes from a lexical item, then the result of the merge operation is $T'_1<T'2$. 
Otherwise, the result is $T'_2>T'_1$. In any case the head of the result comes from $T_1$ and is $\tau'_1 \phi_1$. 
As far as strings are concerned, head-movement is not required: grammars with or without head-movement generate the same languages, the simulation of head-movement requires many standard merge and moves to be simulated, and yields different parse trees. For having shorter derivations and standard parse trees, we prefer to use head movement in this paper. There also exists head movement with left adjunction, which is defined symmetrically the phonological features moving before the phonological features of the selector head. One can also define affix-hopping which is rather similar except that the selectee collects the phonological features of the head of the selector. 

The \emph{move}  rule applies to a single tree $T$ whose head is $+x \tau \phi$ i.e. starts with a movement target $+x\in\place$, such that $T$ has a subtree $T_1$ whose head starts with the corresponding movement trigger  ${-}x\in\move$. Then the result is $T'_1>T'^\circ$ where $T'_à1$ is $T_1$ without the ${-}x$ feature and $T'^\circ$ is $T$ minus the $T_1$ subtree and minus the ${+}x$ feature. Observe that the head of the result is the same as the initial one, except that its first feature ${+}x$  has been erased.\footnote{It is possible to also consider weak movement, which leaves the $-x$ subtree 
 at is pace, but suppress the semantic information, and makes a copy of the tree structure with only the semantic features and moves it as we did. As we do not speak about semantic features, we skipped this possible rule.} 
 
 \begin{defn}{Language generated by a minimalist grammar} 
 A convergent derivation generating the sentence $\phi_1\cdots \phi_n\in \Phi^*$ 
 ends up with a single non phonological feature $\mathsf{C}$ on the head, 
 while the leafs read from left to right  yield $\phi_1\cdots \phi_n\in \Phi^*$. 
 \end{defn} 

An important requirement on minimalist grammars, both in the principles of Chomsky and in Stabler's formalisation,  is the condition known as the \emph{shortest move}. In Chomsky's formulation it says that whenever two subtrees have 
 a ${-}x \tau$ head, the first one, that is the closes to the target, moves. In Stabler's view it says that whenever there are two candidates to movement, i.e. when there are two leaves starting with ${-}x$, the derivation crashes. Unless otherwise stated, with stick to Stabler's definition.

Minimalist grammars have rather interesting formal properties, some of which are worth quoting. 
Indeed, Harkema, Kobele,\linebreak  Michaelis, Morawietz, M\"onnich, Retor\'e, Salvati, Stabler 
studied minimalist grammars from the viewpoint of formal language theory, 
the main focus being their generative capacity as sets of strings and as sets of trees, 
the complexity of parsing and learning strategies. We recall here some of the results: 

\begin{itemize} 
\item The generative capacity of minimalist grammars (with shortest move) is equivalent to that of context free linear rewriting system. \cite{Mic98,Mic01} 
\item In the absence of shortest move, it has been showed that the membership problem is equivalent to the yet unknown decidability of provability of Multiplicative Exponential Linear Logic.  \cite{Salvati07pauillac}
\item 
As far as trees are concerned, we know that the minimalist derived trees, that are the ones depicting syntactic structures can be obtained from a regular tree grammar with a transduction applied thereafter. 
\cite{MMM00,KRS07}
\end{itemize} 

Although there are some semantic concerns in minimalist grammars as rewriting systems e.g. in \cite{Kobele06phd} what is a logical form and how it could be computed is not as clear as it is in the logical approach to categorial grammars. This is likely to be the most frustrating limit of minimalist grammars and the main reason for studying them from a categorial perspective. 

For us,  the main advantage of minimalist grammars (or in a broader perspective, generative grammar, transformational grammars, etc.)
is their linguistic richness and soundness, which is attested by the relation they are able to draw between sentences (questions and related declarative sentences for instance) or between languages (the differences among features are a way to explain language variation). 

\section{Categorial grammars}

A categorial grammar consists in a lexicon which associates with every lexical item a finite set of categories (propositional logical formulae).  In the case of AB grammars 
categories are finitely generated by the connectives $\lfrom$ and $\lto$ from a set of base categories: $S$ (sentence), $n\!p$ (or $d\!p$ for noun phrase or determiner phrase), $n$ (common noun) etc. 
A sequence of words $w_1 \cdots  w_n$ is in $L(\Lex)$ whenever 
$\forall i \exists t_i\in \Lex(w_i)$ such that $t_1,\ldots, t_n\seq S$ is derivable in a logical calculus. For AB grammars  the calculus simply consists in the two famous residuation laws, which are elimination rules from the logical viewpoint, i.e. when $\lfrom$ and $\lto$ are viewed as non commutative linear implications. 

Alternatively, one could also label the formulae in the proofs as follows: 

\begin{center} 
\begin{tabular}{cc} 
\multicolumn{2}{c}{\begin{prooftree}
(w: \mbox{word, with $A\in\Lex(w)$ })
\justifies 
\seq w:A 
\using axiom 
\end{prooftree} }
\\ &  \\   &   \\  
\begin{prooftree} 
\seq w_1\ldots w_n: A 
\quad\seq m_1\cdots m_p: A\lto B
\justifies 
\seq w_1\cdots w_n m_1\cdots m_p: B
\using\lto_e 
\end{prooftree} 
\\ &  \\   &  \\ 
\begin{prooftree} 
\seq t: B\lfrom A
\quad 
\seq u: A 
\justifies 
\seq t u: B
\using\lfrom_e 
\end{prooftree} 
\end{tabular} 
\end{center} 

This lead to a definition of the generated language as typable strings: 

\begin{defn}{Language generated by an AB grammar}
\label{Stypable}
(*)  $w_1 \cdots w_n \in L(\Lex)$ 
whenever $\seq w_1 \cdots w_n: S$. 
\end{defn}

The logical calculus we are to use for categorial minimalist grammars is close to the one above since we shall 
consider labels and only elimination rules. But it also resembles the Lambek calculus since we have more logical rules and one of them, 
namely product elimination, requires axioms and variables.\footnote{Can one extends this definition of the generated language  to  Lambek grammars? 
Lambek calculus extends 
the above calculus with the introduction rules (or abstraction),
and this requires to have context and variables
 (introduced by the axiom $x:A\seq x:A$). 
Since variables are abstracted only from leftmost or rightmost positions,  one has to carry over context with words to prohibit illicit abstractions, 
hence the lightness of the above system  without contexts is lost.}

One can wonder why moving out the mathematical paradise provided by usual categorial grammars, 
AB grammars or Lambek grammars, which are very neat and elegant, especially the Lambek calculus.  
The main reason is that the syntactic abilities of categorial grammars are limited. 
A formal way to say so is that they only describe context-free languages, which are commonly assumed to be insufficient to describe natural language syntax,
but a more empirical one is their difficulty to describe various syntactic phenomena, like discontinuous constituent, 
or to relate declarative sentence and questions. 
\footnote{Although Lambek grammars only describe context free languages, if the grammar is not compiled 
into a context free grammar, parsing by proof search is NP complete.}  Therefore, one has to extend such systems, either by extending the logic by modalities and postulates, as \cite{Moo88} did, 
thus sticking to the parsing as deduction paradigm, or to have a logical basis providing the deep structure and an extra mechanism computing word order. 

Despite their syntactic limitations, categorial formalisms nevertheless allow an appealing direct computation of (usual) logical formulae from a syntactic analysis. The deep reason, already noticed by the Ancients is that syntactic categories correspond to semantic types, see e.g. 
\cite{BD81}: a sentence $S$ is a proposition $t$, 
a definite $np$ is an individual $e$, a common noun $n$ or an intransitive verb $np\lto S$ is a one place predicate, a transitive verb is a two place predicate etc. 
Without providing the details, since we are going to present a similar process for categorial minimalist grammars, let us explain how it works.

In order to compute the semantic formula associated with a sentence, we have:  
\begin{itemize}
\item A lexicon that provides each word with a syntactic category $c$ and a lambda term whose type is the semantic translation $t=c^*$. 
\item A syntactic analysis that is a proof in the Lambek calculus of $S$ under the assumption $t_1,\ldots,t_n$, each $t_i$ being a possible category for the word $w_i$. 
\end{itemize} 
The algorithm which computes the associated logical formula is quite simple: 
\begin{itemize}
\item[(i)] Convert the syntactic categories into semantic types as follows: $np^*=e$, $S^*=t$, $n^*=e\fl t$, $(A\lto B)^*=(B\lfrom A)^*=e\fl t$: this yields a proof in intuitionistic logic of $S^*=t$ under the assumption $c_1^*=t_1,\ldots,c_n^*=t_n^*$. This proof is a proof that the lambda term hereby defined corresponds to a formulae (is of type $t$), whose free variables are of type $t_i$.
\item[(ii)] Replacing each variable with the corresponding term of type $t_i$ provided by the lexicon provide a close term of type $t$, that is a logical formulae. 
\end{itemize}

\section{Categorial minimalist grammars} 

\subsection{Labels encoding word order} 

Let us consider a set of variables $V$ (for expressing hypothesis and encoding movement) and a (disjoint) set of phonological forms $\Phi$ (words, lexical items) 
The labels we are to use are slightly more complicated than sequences of words. A label $\vec r$ is a triple each component being a sequence of variables and phonological forms 
 $$\vec r=(r_{spec}\,|\, r_{head}\,|\, r_{comp})=(r_{s}\,|\, r_{h}\,|\, r_{c}) \in \left[(V\cup \Phi)^*\right]^3$$

Intuitively, these three strings are the yields of the three subtrees respectively corresponding to the head ($r_h$), the specifier ($r_s$) and the complement
($r_c$). The following notations are convenient to denote the suppression of one of the components: 
$\vec r_{-h} = (r_{s}\,|\, \epsilon\,|\, r_{c})\in \left[(V\cup \Phi)^*\right]^3$, 
$\vec r_{-s} = (\epsilon\,|\, r_{h}\,|\, r_{c})\in \left[(V\cup \Phi)^*\right]^3$, 
$\vec r_{-c} = (r_{s}\,|\, r_{h}\,|\, \epsilon)\in \left[(V\cup \Phi)^*\right]^3$. 
With these notations $r_{-h}=r_{s}\, r_{c}\in (V\cup \Phi)^*$

Given a label $\vec r=(r_{s}\,|\, r_{h}\,|\, r_{c})$ we denote by $r$ the concatenation of its three components: $$r= r_{s}\, r_{h}\, r_{c}\in (V\cup \Phi)^*$$

\subsection{A restricted fragment of partially commutative logic}

To represent minimalist grammars we make use of a very restricted fragment of partially commutative linear logic. This later system introduced by de~Groote in \cite{Gro96} and extended in \cite{Ret04petri},  is a superimposition of the Lambek calculus (intuitionistic non commutative multiplicative linear logic) and intuitionistic commutative multiplicative linear logic. The connectives are both  the ones of the Lambek calculus ($\bullet, \lto,\lfrom$) and the ones of (commutative)multiplicative linear logic ($\otimes, \multimap$).

We only consider a particular formulation of a restricted fragment of the calculus to encode minimalist grammars: 
\begin{itemize} 
\item We use natural deduction as in \cite{AR07cie}. 
\item 
We only use the commutative conjunction ($\llts$) and the two non commutative implications ($\lfrom,\lto$).   
\item We only use elimination rules. 
\item Contexts will simply be multisets of formulae.   
Usually, for such calculi, contexts are partially ordered multisets of formulae and a rule called \emph{entropy} allows to relax such order. 
In the restricted calculus we use, we systematically use the full strength of \emph{entropy}
just  after the eliminations of $\lfrom$ and $\lto$ which usually introduce order between the assumptions. Hence contexts are simply multisets of formulae. 
\end{itemize} 

The product elimination rule is especially important, because it enables a coding of movement (or rather \emph{copy theory} of \cite{Bro95}) from one hypothesis to the other one.  
It is a free adaptation of the product elimination rule to the proof expressions  of \cite{Abr93cill} (see figure \ref{productelim}).

\begin{figure}
\begin{center} 
\prooftree
	\Delta \vdash  z:A \otimes B
	\quad \Gamma[(x:A,y:B)]\vdash  t:C
	\justifies
	\Gamma[\Delta]  \vdash  \left(\mbox{let}\ (x,y)=z\ \mbox{in}\ t\right):  C 
	\using
             [\otimes_{e}]
         \thickness=0.07em
\endprooftree
\end{center} 
\caption{Labels for product elimination rule.} 
\label{productelim}
\end{figure}

The formulae we use have a definition that follows rather precisely the sequences of features that are used in minimalist grammars, i.e. in the set $\mgentries$ defined in \ref{mgentries}: 

\begin{defn}{Logical formulae in a categorial minimalist lexicon}
$$
\begin{array}{lcl}
\formcmg &::=& \mathsf{x} \lfrom \mathsf{b} \ |\  \mathsf{x} \lfromhdr \mathsf{b} \ |\  \mathsf{c}\\
\mathsf{x} &::=& \mathsf{b} \lto \mathsf{x}  \ |\  \mathsf{b} \ltohdr \mathsf{x}  \ |\  \mathsf{m} \lto \mathsf{x} \ |\  \mathsf{c}\\
\mathsf{c} &::=& \mathsf{m} \otimes \mathsf{c} \ |\  \mathsf{b} \\
\end{array}
$$
\label{cmgformulae} 
\end{defn} 

The rules which control the labelling are presented in figure \ref{pcmllrules} --- observe that labels do not control the deductive process but are derived from deductive process. 

\begin{figure} 
\begin{center}
\prooftree
	\langle s, A \rangle \in Lex
	\justifies
	\seq_G (\epsilon\,|\, s\,|\, \epsilon) : A
	\using [Lex]
\endprooftree

\bigskip

\prooftree
	x \in V
	\justifies
	x: A \seq_G (\epsilon\,|\, x\,|\, \epsilon) : A
	\using [axiome]
\endprooftree

\bigskip

\prooftree
	\Gamma \seq_G \vec r_1 : A \lfrom B
	\quad \Delta \seq_G \vec r_2 : B
	\justifies
	\langle \Gamma ; \Delta\rangle \seq_G (r_{1s}\,|\,r_{1t}\,|\,r_{1c}\, r_2) : A
	\using [\lfrom_e]
\endprooftree

\bigskip

\prooftree
	\Delta \seq_G \vec r_2 : B
	\quad \Gamma \seq_G \vec r_1 : B \lto A
	\justifies
	\langle \Gamma ; \Delta\rangle \seq_G  (r_2\, {r_{1}}_s\,|\, {r_{1}}_h\,|\, {r_{1}}_c) : A
	\using [\lto_e]
\endprooftree

\bigskip

\prooftree
	\Gamma \seq_G \vec r_1 : A \otimes B
	\quad \Delta[x: A , y:B] \seq_G \vec r_2 : C
	\justifies
	\Delta[ \Gamma] \seq_G   \vec r_2[r_1/x , \epsilon/y ] : C
	\using [\otimes_e]
\endprooftree

\bigskip

\prooftree
	\Gamma \seq_G \vec r : A
	\quad \Gamma' \sqsubset \Gamma
	\justifies
	\Gamma' \seq_G \vec r : A
	\using [\sqsubset] 
\endprooftree
\end{center}
\label{pcmllrules}
\caption{Labelled PCMLL deductions. As renaming allows, it is assumed that no variable is common to two labels from different premises: $Var(\vec{r}_1)\cap Var(\vec{r}_2)=\emptyset$. In the $\otimes_e$ rule, $A$ is usually a movement feature.}
\end{figure}

The rules of figure  \ref{pcmllrules}  will be used as a group of rules which corresponds to 
the main operations of minimalist grammars, namely \emph{merge} and \emph{move} given in (\ref{merge}) and (\ref{move}).

We also need to distinguish different  $\lto,\lfrom$-eliminations because head-movement with right adjunction
is as far as logical types are concerned a \emph{merge} but is has a different effect on word order. 
The two rules corresponding to lexical merge / head-movement with right adjunction $\lfromhdr$ or the non lexical variant are provided in figure \ref{hdrrules} 
This is the reason why we use the symbols $\ltohdr$ and $\lfromhdr$ corresponding to head-movement with right adjunction. 
Clearly, this is a weakness of the system, because as far as proofs are concerned it is weird to have 
different connectives with the same rules. 
We can answer to the possible protests that:
\begin{itemize}
\item As for minimalist grammars, head-movement is not really required since it can be simulated with regular movement, hence \emph{merge} and \emph{move} which are completely directed by the formulae Ñ 
\item Because the system does not contain the introduction rules, the system is unable to prove $A\lto B\equiv A\ltohdr B$ nor $A\lfrom B\equiv A\lfromhdr B$ which would be quite a problem. 
\end{itemize}

\begin{defn}{The operation \emph{merge} in a categorial setting}
\begin{center}
\prooftree
	\langle sem, A \rangle \in Lex
	\justifies
	\seq_sem : A
	\using [Lex]
\endprooftree

\bigskip

\prooftree
	\seq g : A\lfrom B
	\quad \Delta \seq u : B
	\justifies
	\Delta \seq (g(u)) : A
	\using
	  [mg]
\endprooftree

\bigskip 

\prooftree
	 \Delta \seq u : B
	\quad \Gamma \seq g : B \lto A
	\justifies
	\Delta , \Gamma \seq (g(u)): A
	\using
	  [mg]
\endprooftree
\label{merge} 
\end{center} 
\end{defn} 

\begin{defn}{The operation  \emph{move} in categorial setting}
\begin{center} \prooftree
	\quad \Gamma \seq \vec s : A \otimes B
	\quad  \Delta[u : A, v: B] \seq \vec r : C
	\justifies
	 \Delta[\Gamma] \seq \vec r[u:=s , v:=\epsilon] : C
	\using
	  [mv]
\endprooftree
\end{center} 
\label{move} 
\end{defn} 

\begin{defn}{Head-movement with right adjunction in a categorial setting}
\begin{center}
\prooftree
	 \Gamma \seq (r_{s}\,|\, r_{h}\,|\, r_{c}) : A\lfromhdr B
	\quad \quad \Delta \seq (s_{s}\,|\, s_{h}\,|\, s_{c}) : B
	\justifies
	\Gamma , \Delta \seq (r_{s}\,|\, r_{h} \, s_{h}\,|\, r_{c}\, s_s\, s_c) :  A
	\using
	  [mg hdr]
\endprooftree

\vspace{2ex}

\prooftree
	\Delta \seq   (s_{s}\,|\, s_{h}\,|\, s_{c}) : B
	\quad \quad \Gamma \seq (r_{s}\,|\, r_{h}\,|\, r_{c}) : B \ltohdr A
	\justifies
	\Delta , \Gamma \seq (s_{s}\, s_c \, r_{s}\,|\,  r_{h} \, s_{h} \,|\, r_{c}) : A
	\using
	  [mg hdr]
\endprooftree
\end{center}
\label{hdrrules} 
\end{defn}

\subsection{Grammars} As earlier said, the  definition of the generated language 
resembles the one of categorial grammar given in \ref{Stypable}, although here it encompasses a broader notion of label. 

\begin{defn}{Language generated by a categorial minimalist grammars}
A grammar produces a sentence $\phi_1 \ldots \phi_n\in\Phi^*$ whenever there exists a derivation 
of $\vdash \vec r: C$ with $r=\phi_1 \ldots \phi_n$ (remember that $r$ is the concatenation of the three components of $\vec r$). 
This definition entails among others that variables must disappear with product elimination rules encoding movement during the derivation process. 
\end{defn} 

A natural question is whether minimalist grammars and categorial grammars generate the same strings and derived trees.
The answer is yes,  and the  proof may be found in \cite{Amblard07phd},\footnote{Actually \cite{Amblard07phd} does not include the head movemen case, but especially for this construct, the two kind grammars work just the same, \emph{mutatis mutandi}.}. It relies on translations from one king of lexicon to the other:

\begin{defn}{The translation $(\_)^\alpha$ from a lexicon \`a la Stabler into a categorial minimalist lexicon} 
\begin{center}
$\begin{array}{lll}
(\gamma\; {-}f)^{\alpha} &=& (f \otimes (\gamma)^{\delta})\\
({=}f \; \gamma)^{\alpha} &=& (\gamma)^{\beta} \lfrom f\\
( f {\hdr} \gamma )^{\alpha} &= &(\gamma)^{\beta} \lfromhdr  f\\ 
(f)^{\alpha} &=&  f\\ 
({=}f \;\gamma)^{\beta} &=& f \lto (\gamma)^{\beta}\\
( f {\hdr}\; \gamma)^{\alpha} &=& f \ltohdr (\gamma)^{\beta}\\Ê
({+}f \;\gamma)^{\beta} &=& f \lto (\gamma)^{\beta}\\
(\gamma \; {-}f)^{\beta} &=& (f \otimes (\gamma)^{\delta})\\
(f)^{\beta} &=& f\\
(\gamma\; {-}f )^{\delta} &=& (f \otimes (\gamma)^{\delta})\\
(f)^{\delta} &=& f
\end{array}$
\end{center}
\label{mgtocmg} 
\end{defn}

\begin{defn}{The translation $(\_)^{\alpha'}$ from  a categorial minimalist lexicon into a	 lexicon \`a la Stabler}
\begin{center}
$\begin{array}{lll}
(F \lfrom f)^{\alpha'} &=& {=}f \; (F)^{\beta'} \\
(F \lfromhdr f)^{\alpha '} &=& f{\hdr} \; (F)^{\beta '}\\ 
(f \otimes F)^{\alpha'} &=& (F)^{\delta'} \; {-}f \\
(f)^{\alpha'} &=& f\\
( f \lto F)^{\beta'} &=& 
\left \{ \begin{array}{l}
{=}f \;(F)^{\beta'} si\; f \in $\textsc{$p_1$}$\\
{+}f \;(F)^{\beta'} si\; f \in $\textsc{$p_2$}$ 
\end{array}\right .\\
(f \ltohdr F )^{\beta '} &= &f\hdr  \; (F)^{\beta '}\\Ê
( f)^{\beta'}  &=& f\\
(f \otimes F)^{\beta'} &=& (F)^{\delta'}\; {-}f\\
(f \otimes F)^{\delta'}  &=& (F)^{\delta'}\; {-}f\\
(f)^{\delta'}  &=& f 
\end{array}$
\end{center}
\label{cmgtomg} 
\end{defn}

\begin{defn}{Equivalence of categorial minimalist grammars and minimalist grammars} 
If a minimalist grammar \`a la Stabler is turned into a categorial minimalist lexicon according to the rules in figure \ref{mgtocmg},
then the generated sentences are the same and even parse trees are almost the same (provided the subtree undergoing movement is correctly placed). Conversely, if one turns a  categorial minimalist lexicon (only containing formulae in $CMG$ defined in figure \ref{cmgformulae}) into a minimalist lexicon according to the rules in \ref{cmgtomg},
then one also gets the same generated sentences and parse trees. 
\end{defn}

A remaining question is how constraints on derivations, like \emph{shortest move}, could be formulated in a deductive setting. This is a weakness of the system: the only way to formulate these restrictions is to impose constraints on derivations, and this is not so appealing from a logical viewpoint since rule are usually context free, i.e. apply locally without referring to the history of the deduction --- unless one goes to dependent types, which are complicated and not yet explored for linear calculi.

\section{Syntax and semantics in categorial minimalist grammars} 
\label{Hmorphism} 

The purpose of viewing minimalist grammars as a categorial system is to have an automatic conversion of syntactic compositions into semantic composition. 
We firstly present the semantic types and their relation to syntactic categories, 
then a variant of $\lambda$-DRT with $\mu$ constructs in order to obtain the different scope readings from a single semantic representation, 
and finally sum up the algorithm producing semantic analyses. 

\subsection{From syntactic categories to semantic types}
The formulae that we infer as semantic representations are first order, because we are going to use DRT which better matches 
the discourse structure. 
Therefore we reify predicates, we use variables to express high order properties: we write 
$run(e,Rex) \land fast(e)$ instead of $fast(run(Rex))$. 
Consequently we have the following three base types 
\begin{itemize} 
\item[$\montt$] the standard Montague type, for truth values and propositions
(that will be used as $\bot$ from the $\mu$-calculus viewpoint).  
\item[$\monte$] the standard Montague type, for entities. 
\item[$\montev$] the type for reifying events and predicates in order to  remain in first order logic. 
\end{itemize}

The standard translation $H$  of a syntactic category $\mathsf{x}$ into the corresponding semantic type $H(\mathsf{x})$
has to be slightly extended, because in addition to the syntactic connectives $\lfrom,\lto$ we have $\llts$, together with 
an extra basic type for reifications. 

\begin{defn}{Converting syntactic categories into semantic types}
$$
\begin{array}{|rcl|} \hline 
H(\mathsf{C}) &=& \montt \\ \hline 
H(\mathsf{v}) &=&   \montev \rightarrow \montt  \\ \hline 
H(\mathsf{t}) &=&   \montev \rightarrow \montt  \\ \hline 
H(\mathsf{V}) &=&  \monte \rightarrow  ( \montev \rightarrow \montt ) \\ \hline 
H(\mathsf{d}) &=& \monte \\ \hline 
H(\mathsf{n}) &=& \monte \rightarrow \montt \\ \hline\hline 
H(\mathsf{x\llts y})  & = & H(\mathsf{y})\ \mbox{if} a\in \mathsf{m}\\ 
& = & H(\mathsf{x})\ \mbox{if\ } a\not\in \mathsf{m}\\ \hline 
H(\mathsf{x}\lto \mathsf{y})&=&H(\mathsf{x})\rightarrow H(\mathsf{y})\\ \hline 
H(\mathsf{y}\lfrom \mathsf{x})&=&H(\mathsf{x})\rightarrow H(\mathsf{y})\\ \hline 
H(\mathsf{x}\ltohdr \mathsf{y})&=&H(\mathsf{x})\rightarrow H(\mathsf{y})\\ \hline 
H(\mathsf{y}\lfromhdr\mathsf{x})&=&H(\mathsf{x})\rightarrow H(\mathsf{y})\\ \hline 
\end{array} 
$$
\end{defn} 

This translation can be illustrated on the syntactic categories we are to use in the example of section \ref{completeexample}.  

$$   
\begin{array}{|rcl|} \hline 
H(\mathsf{k\llts d}) &=& \monte\\ \hline 
H(\mathsf{V\lfrom d}) &=&    ( \montev {{\rightarrow}} \montt) {\rightarrow} \monte {\rightarrow}  \montev {\rightarrow} \montt\\ \hline 
H(\mathsf{(k\backslash (d \backslash v))\lfromhdr V})&=&    (\monte {\rightarrow}  \montev {\rightarrow} \montt) {\rightarrow} \monte{\rightarrow} \monte {\rightarrow}  \montev {\rightarrow} \montt  \\ \hline 
H(\mathsf{(k \lto t )\lfromhdr v})&=&    ( \montev \rightarrow \montt) \rightarrow \monte \rightarrow  \montev \rightarrow \montt\\ \hline 
H(\mathsf{k \otimes d \lfrom n}) &=&   (\monte \rightarrow \montt) \rightarrow \monte\cr  \hline 
H(\mathsf{C \lfrom t}) &=&     ( \montev \rightarrow \montt) \rightarrow \montt \cr 
\hline
\end{array}
$$

\subsection{$\lambda\mu$-calculus} 

Semantics can be computed in standard Montagovian terms and types, as  did \cite{ALR03nancy} or \cite{LR02}
We extend a bit this process, by moving from $\lambda$-calculus to $\lambda\mu$ calculus, a 
$\lambda$ calculus for depicting classical proofs (while standard $\lambda$-calculus correspond to proofs in intuitionistic propositional logic). 
The advantage is that quantifiers do not need to be type raised and that the different scopes  are obtained from a single syntactic semantic analysis,
by different $\lambda\mu$ computations, as initiated by \cite{degroote01mucalcul} and developed for categorial minimalist grammars in 
\cite{Amblard07phd} or \cite{Lecomte08fg}. 

The $\mu$ calculus is strongly related to the formulation of classical logic in natural deduction 
 using \emph{reductio as absurdum}: 
instead of concluding $\lnot\lnot A= (A\rightarrow \bot) \rightarrow \bot$ from $\bot$ under the assumption $\lnot A$ one concludes $A$. Provided one does not use the $\bot$ rule (\emph{ex falso quod libet sequitur}), 
this can be done with any constant type for $\bot$.  Here, as in \cite{degroote01mucalcul}, we shall use Montague $\montt$ type (truth values, propositions) for the $\bot$ of the glue language that is the $\lambda\mu$-calculus. As opposed to common intuitions,  this is really harmless!

We will use a restricted calculus, whose typing rules are very easy:
basic types are the aforementioned $\montt,\monte,\montev$ 
and, as said above,  $\montt$ will be used as $\bot$. 
Types are defined as usual, with the arrow (intuitionistic implication) only.

For each type $Z$, we have a denumerable set of $\lambda$-variables of type $Z$. 

We have $\mu$ variables with type $X \rightarrow\montt$ for some type $X$.  

There are three kinds of terms, Values $V$, Unamed Terms $v$ and Named Terms $c$ with $V\subset v$,
which are defined as the smallest sets closed under the following operations: 
\begin{itemize} 
\item 
a $\lambda$-variable $x:X$ is a value and an unnamed  term.  
\item 
if $u:U$ is an unnamed term and $x$ a $\lambda$ variable  then $\lambda x.\ u: X\rightarrow U$ is a value and an unnamed term
\item 
if $v_1:X\rightarrow U$ and $v_2:X$ are unnamed terms, $(v_1(v_2)):U$ is an unnamed term 
\item 
if $v:X$ is an unnamed term and $\alpha:X\rightarrow\montt$ a $\mu$-variables, than $(\alpha (v)):X$ is a named term
\item 
if $v:X$ is a named term and $\alpha:X\rightarrow\montt$ is a $\mu$-variable then $(\mu\alpha.\ v):X$ is an unnamed term. 
\item 
Any term of type $\montt$ is a named term (this follows from $\bot=\montt$). 
\end{itemize}

The reduction patterns for $\lambda\mu$-calculus include the usual $\beta$ conversion as well 
as reductions concerning $\mu$ variables and the $\mu$-binder: 

\begin{defn}{$\lambda\mu$ reductions}
\label{mureduc} 
\begin{center}
$
\begin{array}{lllll}
(\beta) & ((\lambda x.\ t) (u)) &\rightsquigarrow_{\beta} & t[u/x]\\Ê
(\mu)&((\mu \alpha .\ T[(\alpha\ Q)]) P) &\rightsquigarrow_{\mu} &\mu \alpha .\ T[(\alpha (Q\ P))]\\
(\mu')&(P (\mu \alpha .\ T[(\alpha\ Q)])) &\rightsquigarrow_{\mu'} &\mu \alpha .\ T[(\alpha (P\ Q))]\\
(\varsigma)&((\mu \alpha.\ T[(\alpha\ Q)]) &\rightsquigarrow_{\simpl} &T[Q] \qquad \mbox{ only for }\ Q:\montt \\
\end{array}
$
\end{center}
\end{defn} 

Let us recall that $\lambda\mu$ calculus, with the given reductions,  is \emph{not confluent}. This will be useful to obtain the many readings of a given sentence, as first observed by \cite{degroote01mucalcul}. 

\subsection{$\lambda\mu$-DRS} 

Instead of standard Montague semantics, 
we prefer to produce Discourse Representation Structures (DRS), whose dynamics better matches the one of sentences and discourse. 
Of course, given the wished correspondence with the categorial framework, we use a compositional presentation of DRT, 
like $\lambda$-DRT.  But, because we also wish to have $\mu$ reductions for avoiding type raising and solving scope questions, we use 
a generalisation of $\lambda$-DRT  that we call $\lambda\mu$-DRT. 

Basically, we consider DRS with $\lambda$ and $\mu$ variables 
(all $\mu$ variables will be of type $X\rightarrow \montt$),  using freely both $\lambda$ and $\mu$  abstractions,
and we refer to these structures as $\lambda\mu$-DRS (Discourse Representation Structure) 
In addition to the  $\lambda$ and $\mu$ variables  and the variables used for movement which can be considered as free $\lambda$ variables 
we need a set of discourse refrents $V_D$ of type $\monte$ (some could be of type $\montev$) 
denoted by $d_1, \cdots, d_n$.

%
%

The lexicon in addition to the syntactic categories maps each lexical item to a $\lambda\mu$-DRS. 
A $\lambda\mu$-DRS is typed term with three different kinds of variables $V_\lambda$ (containing movement variable, which have nothing special except that they are not bound by any $\lambda$), 
$V_\mu$ and $V_D$ and three corresponding kinds of binders the later one allowing dynamic binding. Variables that are $\mu$ or $\lambda$ bound can be suppressed from the context, while  discourse referents of $V_D$ should remain in the context until some simplifications are performed, as their binding is sophisticated, see \ref{intern}. 

Types for these terms are defined as usual, from three base types $B=\{\monte, \montev, \montt\}$ (entities, events, truth-values) by the arrow: 

$$T= B\ |\ T\rightarrow T$$

We have constants, namely predicates of type $\upsilon \rightarrow (\upsilon \rightarrow  (\upsilon \rightarrow \cdots \montt))$ where $\upsilon$ is either $\monte$ or $ \montev$ ---  there is no need to systematically force that a predicate only involves one event, itself. 
As usual, an intransitive verb has type $ \montev\rightarrow\monte\rightarrow \montt$, 
a transitive verbe $ \montev\rightarrow\monte\rightarrow \monte\rightarrow \montt$,  
a ditransitive verbe $ \montev\rightarrow\monte\rightarrow \monte\rightarrow\monte\rightarrow  \montt$,  
and a property $\monte\rightarrow \montt$. 

We also have logical constants: 
\begin{itemize} 
\item ${\land}:\montt \rightarrow \montt \rightarrow \montt$ 
\item ${\dland}:\montt \rightarrow \montt \rightarrow \montt$ 
\item ${\Rightarrow}:\montt \rightarrow \montt \rightarrow \montt$
\item $\dot{=}: \monte \rightarrow \monte \rightarrow \montt$
\end{itemize} 

There is no constant for $\exists$ ($[d|F]$ is close to $\exists d\ F$) nor for $\forall$ ($[|[d|F]\Rightarrow [|G]]$ is close to $\forall d\ (F\Rightarrow G)$), since we thus obtain better scope properties. 
The variant of $\land$ named $\dland$ has a slightly different behaviour with respect to discourse variables. 
The operation $\dland$ concerns DRS as well as formulae,   and is a kind of a fusion, similar to the merge operations explored by \cite{musk:comb96}.

$$
\begin{array}{|ccc|}
\hline 
\mbox{Variable} & \mbox{Term} & \mbox{Result} \\ \hline \hline 
\alpha:(X\rightarrow t) \in V_\mu & u:X & \mu\alpha.\ u: X \\ \hline 
x:X\in V_\lambda & u:U & (\lambda x. u): X\rightarrow U\\ \hline 
\left.
\begin{array}[c]{c}
d:\monte \in V_d\\ 
d: \montev\in V_d\\ 
\end{array}
\right\}
& u:\montt & {[d|u]}:t\\  \hline 
\end{array} 
$$ 

We also have applications: 
$$\begin{array}{|ccc|} \hline 
\mbox{First Term} & 
\mbox{Second Term} & 
\mbox{Application Result} \\  \hline \hline 
\alpha:(X \rightarrow\montt) & u: X & (\alpha (u)):\montt\\  \hline 
u:X\rightarrow Y & v: X & (u(v)):Y\\ \hline 
\end{array} 
$$

Given a formula $F:\montt$ we use the following shorthands: 
\begin{itemize} 
\item 
let $F:\montt$ then $[|F]=F$ in the very same context. In particular $F\rightarrow G$ can be viewed as $[|F]\Rightarrow [|G]$. 
\item 
let $F:\montt$ then $[d_1|[d_2[d_3|[\cdots [d_n[|F]]]]]]=[d_1\cdots d_n|F]$  the  context being unchanged. 
\end{itemize}

A $\lambda\mu$-DRS is a typed term of this calculus. When it is of type $\montt$, without $\mu$-binder nor $\mu$-variable, 
it is an ordinary $\lambda$ DRS, 
and when, furthermore,  there is neither $\lambda$ binder nor $\lambda$ variables, it is an ordinary DRS (provided formulae involved in implications are viewed as short hands for DRS without discourse variables).
In order to formulate the reduction of $\dland$ which encodes dynamic binding, we need some standard notions on DRS and subDRSs. 
 
A sub DRS is simply a subterm. A subDRS $G$ of a DRS $H$  is said to be a \emph{positive} (resp. \emph{negative}) subDRS of $H$ written $\precp$ (resp. $\precm$) whenever:
\begin{itemize}
\item $G=H$
\item $H=[d|K]$ and $G\precp K$  (resp. $G\precm K$)). 
\item $H=K\land L$ or $H=K\dland L$ and $G\precp K$  (resp. $G\precm K$)
\item $H=K\Rightarrow  L$  and  $G\precp K$  or $G\precm L$  (resp. $G\precp L$ or $G\precm K$) 
\end{itemize} 

The reference markers associated to a DRS $[d_1\cdots d_n|F]$ are the discourse refrents $d_1\cdots d_n$. 
The  accessibility relation is the closure of $({\prec})\cup({\ante})$ where $F\ante G$ whenever $F\rightarrow G$. 
The accessible referent markers of a DRS are the referent markers of the accessible DRS.

As said above,  $\dland$ operation is close to the  DRS merge in the literature, but here it will be called \emph{fusion} because we use \emph{merge} as a term denoting a syntactic operation. Although  it could be defined on the run, the reduction of $\dland$ 
is easier to define on a normal $\lambda\mu$-DRS, 
that is a DRS without $\lambda$ abstraction nor $\mu$ abstractions. It can be reduced when it 
applies to a DRS $D$ and a single formula $F$  with free discourse variables $\fv(F)$.
To view this reduction  as a  fusion (merge), 
firstly turn $F$ into $[\fv(F)|F]$: then the resemblance should be clear. 

\begin{defn}{Internalisation}\label{intern} 
If there exist one largest $K=[k_1\cdots k_s|G]$ positive subDRS of $D$ whose accessible referent 
markers includes $\fv(F)$ then $D\dland F$ reduces to $D$ with subDRS $K$ replaced with\linebreak  $K=[k_1\cdots k_s|G \land F]$. 
If there is no such positive subDRS, or if there is no largest one,  no internalisation is possible. 
\end{defn}  

\subsection{The process of syntactic and semantic derivation}

Assume we have a lexicon which maps a word to its syntactic category $x$ (that can be inferred from a minimalist grammar)
together with the corresponding $\lambda\mu$-DRS with the semantic type $H(x)$ defined in 
\ref{Hmorphism}. Firstly, we should provide the semantic counter part of the syntactic rules. 
Merge does, as usual, correspond to $\lambda$ application: the head is the semantic function which is applied to the semantics of the argument. Head movement  with right adjunction, is just a merge from the semantic viewpoint --- indeed 
the resource consumption is just the same, only word order is different from standard merge. 

\begin{defn}{Semantic counterpart of \emph{merge} and of \emph{head-movement}} 
In the \emph{merge} and \emph{hdr} rules of figures \ref{merge} and \ref{hdrrules} 
let $H(s):H(B)$  be the semantic term corresponding to $\vec s : B$ i
and let $H(r):H(A)\rightarrow H(B)$ be the one corresponding to $\vec r:A\lfrom B$ (or $A\lfromhdr B$ or $B\lto A$ or $B\ltohdr A$) 
  (all the possible syntactic categories for $\vec r$ translated  into the semantic type $H(A)\rightarrow H(B)$). 
  Then  the semantic term associated with 
 the resulting category $A$ is $H(r) (H(s)): H(A)$ --- the functional application of the semantic term associated with the function-category to the 
 semantic term associated with the argument. 
 \end{defn} 

The semantic counterpart of the \emph{move} rule of figure \ref{move} is more complex. As may be observed in the syntactic rule,  
\emph{move} connects two places of respective categories $A$ and $B$, in an elimination rule with main premise $A\otimes B$. 
These two places corresponds to two $\lambda$-variables $u$ and $v$ that by construction also appear in the semantic term. 
Items that can undergo movement, typically 
determiner phrases missing a case, or \emph{wh} constituents, 
have a semantic term $s$ associated with a particular discourse variable $d(s)$ introduced in the lexicon within the semantic term  of the determiner.
\emph{Move} consists in replacing the first variable $u$ with the semantic term $s$, and the second, $v$ with the corresponding discourse variable $d(s)$. 
The internalisation process described above in \ref{intern} is precisely made to connect the formula in which $d(s)$ appears and the DRS 
that derives from $s$ and 
which is meant to bound the discourse referent $d(s)$.

\begin{defn}{The semantic counterpart of \emph{move}} 

In the rule below $d(s)$ is the distinguished discourse variable associated with the movable semantic term~$s$. 

$$
 \prooftree
 \Gamma \seq H(s) {:} A \otimes B
	\quad  \Delta[u {:} H(A), v{:} H(B)] \seq H(r)[u,v] {:} H(C)
	\justifies
	 \Delta[\Gamma] \seq H(r)[u{:=}H(s) , v{:=}d(s)] {:} H(C)
	\using
	  [mv]
\endprooftree
$$

\label{semmove} 
\end{defn}

To sum up, how do we proceed to compute both the syntactic and the semantic structure associated to a sentence $\phi_1\cdots\phi_n\in\Phi^*$? 

\begin{defn}{Computing the parse structure and the logical forms in a categorial minimalist grammar} 
\begin{itemize} 
\item[(i)] Firstly, we construct a proof with the rules $mv$, $mg$ and $hdr$ of $\vec r:\mathsf{C}$ with $r=\phi_1\cdots\phi_n$. 
\item[(ii)] Afterwards or in parallel steps, this provides a $\lambda\mu$-DRS of type $t$ without free $\lambda$ variables, nor with free $\mu$ variables, that is nearly what we were looking for: 
we only have to internalise the formulae concerning some discourse variable $d$ that lie outside the corresponding box which bounds $d$. 
This may seem problematic because it turns some free variables into bounded ones but as we proceed just after the derivation, before any renaming takes place, this is harmless. Also by construction, there cannot be conflict about in which box the formula must be moved. $$[d,f| P(d,f)] \land Q(f) \ \rightsquigarrow [d,f| P(d,f) \land Q(f)]$$ 
Hence we end up with a $\lambda\mu$ DRS $D$ of type $t$ without free variables of any kind,  and we may call $D$ the semantic representation of the sentence $[\![\phi]\!]$. 
\item[(iii)] Nevertheless $[\![\phi]\!]$ still contains $\mu$ variables and abstractions, which prevents us from interpreting directly 
$[\![\phi]\!]$ as a formulae. Performing $\beta$ and $\mu$ reductions ($\mu,\mu', \simpl$) yields ordinary DRS that are usual the semantic representations of the sentence. The different scope readings result from the non confluence of $\mu$ reductions, as in \cite{degroote01mucalcul}. 
\end{itemize} 
\end{defn} 

The next section provides detail on all these steps by running a complete example. 

\section{A complete example of a syntactic and semantic analysis} 
\label{completeexample} 

Here is a samble lexicon from which we will derive the two reading of the ambiguous sentence: \emph{the children ate a pizza}. 
$$
\begin{array}{|l|l|l|}\hline 
\mbox{\bf lexical}  & \multicolumn{1}{|c|}{ \mbox{\bf $\lambda\mu${-}DRS \& semantic type}}& \mbox{\bf syntactic}  \\ 
\mbox{\bf item}    &  & \mbox{\bf category}  \\ 
\hline
\les & \lambda Q. \mu \delta . [|[d| (Q\ d)] \Rightarrow [(\delta\ d)]] & \\ &  ( \monte \rightarrow  \montt ) \rightarrow  \monte &  \sf k \otimes d \lfrom n\\ 
& \mbox{specific discourse variable:}\ d & \\ 
\hline

\un &\lambda Q. \mu \gamma . [p | (Q\ p) \land (\gamma\ p)] & \\ &  ( \monte \rightarrow  \montt ) \rightarrow  \monte  & \sf  (k \otimes d)\lfrom n\\ & \mbox{specific discourse variable:}\ p & \\ 
\hline 

\children & \lambda z . (\child\ z)& \\ &     \monte \rightarrow  \montt &  \sf   n\\ \hline 

\pizza &\lambda z . (\pizz\ z)& \\ &     \monte \rightarrow  \montt  &  \sf   n\\ \hline 

\ate & \lambda x \lambda y \lambda e . \eat(e,x,y)& \\ &    \monte \rightarrow  \monte \rightarrow  \montev \rightarrow  \montt &  \sf  V \lfrom d\\ \hline 

\modif & \lambda R \lambda x_2 \lambda y \lambda e . &\\ 

& R (y,e) \dland \patient(e,u)& \\ &     ( \monte\rightarrow  \montev \rightarrow  \montt ) \rightarrow  \monte \rightarrow  \monte \rightarrow  \montev \rightarrow  \montt &   \sf (k\backslash (d \backslash v))\lfromhdr V\\ \hline 

\infl & \lambda Q \lambda y_2 \lambda e . Q(e)&\\ 
& \land \past(e) \dland \agent(e, y_2)& \\ &     ( \montev \rightarrow  \montt ) \rightarrow  \monte \rightarrow  \montev \rightarrow  \montt &\sf  (k \lto t )\lfromhdr v\\ \hline 

\comp

 & \lambda Q . [e|(Q(e))]& \\ &     ( \montev \rightarrow  \montt  ) \rightarrow  \montt & \sf  C \lfrom t \\
\hline
\end{array}
$$

In order to complete both the syntactic and the semantic process, described in the previous section on this example, 
we are to use the following structure: 

$$
\labitem
{
\begin{tabular}{l}
\mbox{declaration of}\\ 
\mbox{the syntactic}\\ 
\mbox{variables}\\ 
\\ 
\end{tabular} 
}
{
\begin{tabular}{l}
\mbox{declaration of the}\\ 
\mbox{corresponding}\\ 
\mbox{semantic variables}\\ 
\end{tabular} 
}
{\mbox{label, triple of phonological forms}}
{\mbox{syntactic category}}
{\mbox{semantic type}}
{\mbox{DRS}}
$$

\begin{figure} 
\label{dps} 
\derivelem\hhh\iii\kkk{mg} 

\derivelem\sss\ttt\uuu{mg}Ê
\caption{Constructing the two $dp$'s: Two steps of the derivation may be performed independently. They both consist in constructing a determiner phrase by a merge rule and semantically the determiner (some or all) applies to the predicate. Notice that the semantic of quantifiers include a $\mu$-abstraction.}  
\end{figure} 

The first step is a lexical merge with a variable that will be used later on for movement. 
On the semantic side merge corresponds to the application of the verbal lambda term to the variable.

\derivelem\aaa\bbb\ccc{mg}

Next step is a head-movement with right adjunction triggered by $\modif$. Semantically it is also an application,
which introduces the $\patient$ thematic role: 

\derivelem\ddd\ccc\eee{hdr} 

Next step introduces in a merge rule a variable for case that will be used for movement. On the semantic side, the variable gets the patient role. 

\derivelem\fff\eee\gggg{mg} 

A move inserts the $dp$ $\un\,\pizza$ (computed in figure \ref{dps}) into the main derivation. Syntactically, the movement checks the case, and the phonological string replaces the variable. 
Semantically the $\lambda\mu$-DRS replaces the variable $\underline{u}$, and the specific discourse referent $p$ replaces $\underline{v}$.  

\derivelem\kkk{\hspace*{-1ex}\gggg}\llll{mv}

Next a variable corresponding to the subject (another \emph{dp}) is introduced in a non-lexical merge-rule. Semantically, we just apply a $\lambda$-expression to the variable: 

\derivelem\mmm{\hspace*{-2ex}\llll}\nnn{mg}

The verb is now ready to receive its inflection. Syntactically it is a head movement with right-adjunction that glues the inflection mark to the right of the verb 
(\emph{eat}\_$infl$ should be understood as \emph{ate}). Semantically, the $\lambda$-term associated with the inflection is applied to the term we derived so far. 

\derivelem{\hspace*{-3ex}\ooo}{\hspace*{-4ex}\nnn}\ppp{hdr}

A variable is introduced to enable the subsequent movement of the subject. Semantically, the application assigns the agent role to the subject. 

\derivelem{\hspace*{-2ex}\qqq}{\hspace*{-2ex}\ppp}\rrr{mg} 

The subject that is the $dp$ we derived in figure \ref{dps}, is inserted in the main derivation by a move. Replacing the variables attracts the subject to the left most position, and on the semantic side, the $\lambda\mu$-DRS associated with the $dp$ replaces the variable $\underline{w}$, while the other semantic variable $\underline{y}$ is replaced with the specific discourse referent $d$. 

\derivelem\uuu{\hspace*{-5ex}\rrr}\vvv{mv} 

Finally, the complementiser turns the whole sentence into a\linebreak  complement by a merge, and semantically reifies the whole formula. 

\derivelem{\hspace*{-2ex}\xxx}{\hspace*{-5ex}\vvv}\yyy{mg}

Thus the syntactic derivation yields the following $\lambda\mu$-DRS:

$[e|\eat(e,\mu \gamma. [p | \pizz(p) \land \gamma(p)],  \mu \delta . [|[d| \child(d)] {\Rightarrow} [|(\delta\ d)]])\\ \hspace*{1ex} \hfill  {\dland} \patient(e,p) {\dland} \past(e) {\dland} \agent(e, d)]$

Yet we need to internalise the predicates, as explained in \ref{intern}, i.e. to move inside the scope of quantifiers the formulae involving these variables
(notice that this is harmless, because we just performed the derivation, and the variables do not have the same name by coincidence nor by renaming) 
and this leads to the following $\lambda\mu$-DRS: 

\begin{center} 
\begin{minipage}{0.7\textwidth}{
$[e|\eat(e,\\ 
\mu \gamma . [p | \pizz(p) \land \gamma(p) {\land} \patient(e,p) ], \\ 
\mu \delta . [|[d| \child(d)] {\Rightarrow} [|\delta(d){\land} \agent(e, d)]])\\ 
{\land} \past(e)]$} 
\end{minipage} 
\end{center} 

For computing this term, we can drop the final $\land \past(e)$ and the $e$ box, since it is not going to interfere 
with the $\mu$ reductions. But we must be slightly cautious 
and write the $\lambda\mu$-DRS in the  $\lambda$ applicative style $(((\eat e) p) d)$ rather than predicate logic, to make $\mu$ reduction more visible. 

$
W=(((\eat\ e)  (\mu \gamma . [p | (\pizz\ p) \land (\gamma\ p) {\land} ((\patient\ e) p) ) ) (\mu \delta  . Y) )
$

with 

$Y= [|[d| (\child\ d)]Ê{\Rightarrow} [|(\delta \ d){\land} ((\agent\ e) d)]]$ 

We now have to reduce acording to the reduction rules in \ref{mureduc} this $\lambda\mu$-term and this will lead to the two readings. The first one is that the children shared a pizza, while the second one says that for every child there is a pizza that he ate.

\begin{itemize} 
\item[(i)]
$\mu'$ reduction with $\alpha=\gamma$ ($\mu$-variable), $V=(\eat\ e)$ and $(\alpha\ T)= (\gamma\ p)$ replaced by 
$(V (\alpha'\ T))= (\gamma'\ ((\eat\ e) p))$
 yields: 

$W\rightsquigarrow(\mu \gamma'. [p | (\pizz\ p) \land (\gamma'\ ((\eat\ e)p)) {\land} ((\patient\ e) p) ) (\mu \delta  . Y) $ 
\item[(ii)] 
A $\mu$ reduction with $\alpha=\gamma'$ ($\mu$-variables), $V=(\mu \delta  . Y)$ and $(\alpha\ T)=(\gamma'\ ((\eat\ e)p))$ replaced 
by\\Ê $((\alpha'\ T) V)=((\gamma''\ ((\eat\ e)p)) (\mu \delta  . Y))$
yields: 

$W\rightsquigarrow(\mu \gamma''. [p | (\pizz\ p) \land ((\gamma''\ (((\eat\ e)p) (\mu \delta  . Y))) {\land} ((\patient\ e) p) )]) $ 

\item[(iii)] 
Next we apply a $\simpl$-rule on $\gamma''$, yielding: 

$W\rightsquigarrow[p | (\pizz\ p) \land \underbrace{(((\eat\ e)p) (\mu \delta  . Y))}_X{\land} ((\patient\ e) p) ] $ 

\item[(iv)] 
The underbraced subterm:

 $X=(((\eat\ e)p) (\mu \delta  . Y))$ is actually: 
 
$(((\eat\ e)p) (\mu \delta . [|[d| (\child\ d)]Ê{\Rightarrow} [|(\delta \ d){\land} ((\agent\ e) d)]]))$
and it can undergo a $\mu'$-reduction with $\alpha=\delta $ ($\mu$-variable)  and $V= ((\eat\ e)p)$ 
in which $(\alpha\ T)=(\delta \ d)$ is replaced with\\  $(\alpha'\ (V\ T))=(\delta'\ '(((\eat\ e)p)d))$, thus yielding: 

$X \rightsquigarrow (\mu \delta '. [|[d| (\child\ d)]Ê{\Rightarrow} [|(\delta '\ (((\eat\ e)p)d)){\land} ((\agent\ e) d)]])$ 

\item[(v)] finally a simplification by the $\simpl$-rule yields: 

$X \rightsquigarrow [|[d| (\child\ d)]Ê{\Rightarrow} [|(((\eat\ e)p)d){\land} ((\agent\ e) d)]]$
\end{itemize} 

If we insert the result of $X$ into the result of $W$ and add the final $\past(e)$ 
we obtain: 
 $[e|[p | (\pizz\ p) \land [|[d| (\child\ d)]Ê{\Rightarrow}\\ \hspace*{1ex} \hfill  [|(((\eat\ e)p)d){\land} ((\agent\ e) d)]] {\land} ((\patient\ e) p) ] \land \past(e)]$

 that is $\exists e. \past(e)\land\\  \exists p \left(\pizz(p) {\land} \patient(e,p) \land \\ \hfill \forall d \left(\child(d) \Rightarrow \left(\eat(e,p,d) \land \agent(e,d)\right)\right) \right)$

As $\lambda\mu$-reduction is not convergent, other readings may be obtained by applying different reductions 
when possible. For instance, after step 1 for which there is no choice, yielding

$W\rightsquigarrow (\mu \gamma'. Z) (\mu \delta  . [|[d| (\child\ d)]Ê{\Rightarrow} [|(\delta \ d){\land} ((\agent\ e) d)]])$ 

with $Z=[p | (\pizz\ p) \land (\gamma'\ ((\eat\ e)p)) {\land} ((\patient\ e) p) ]$

instead of the $\mu$ reduction used in the first derivation, we can apply a $\mu'$ rule. 
 \begin{itemize}
 \item[(ii)]A $\mu'$-rule with $\alpha=\delta $, 
 and $V= (\mu \gamma'. Z)$ $(\alpha\ T)=(\delta \ d)$ replaced with 
 $(\alpha'\ (V\ T))=(\delta' \ ((\mu \gamma'. Z) d))$
 yields:

 $W\rightsquigarrow\mu \delta ' . [|[d| (\child\ d)]Ê{\Rightarrow} [|(\delta '\ ((\mu \gamma'.\ Z) d)){\land} ((\agent\ e) d)]]$ 



\item[(iii)] next we can apply a simplification $\simpl$ rule for $\mu \delta '$ yielding: 

 $W\rightsquigarrow [|[d| (\child\ d)]Ê{\Rightarrow} [|\underbrace{((\mu \gamma'. Z) d)}_S{\land} ((\agent\ e) d)]]$ 

\item the underbraced subterm
$S=((\mu \gamma'. Z) d)$ that is 

$S=((\mu \gamma'.[p | (\pizz\ p) \land (\gamma'\ ((\eat\ e)p)) {\land} ((\patient\ e) p) ]) d)$ 
 
 can undergo a $\mu$-reduction 
with  $\alpha=\gamma'$, $\beta=\gamma''$ ($\mu$-variables) and $V=d$ 
in which  $(\alpha T)= (\gamma'\ ((\eat\ e)p))$ 
replaced with $(\alpha (T V))= (\gamma'\ (((\eat\ e)p)d))$
yielding: 

$S\rightsquigarrow (\mu \gamma''.[p | (\pizz\ p) \land ((\gamma''\ ((\eat\ e)p)) d) {\land} ((\patient\ e) p) ])$

\item[(iv)] next we can apply to $S$ a $\mu \gamma''$ simplification, yielding: 

$S\rightsquigarrow [p | (\pizz\ p) \land (((\eat\ e)p) d) {\land} ((\patient\ e) p) ]$

\end{itemize} 

If we insert the reduct of $S$ into what we reduced $W$ to, we get: 

\noindent 
$W\rightsquigarrow\\ {}  [|[d| (\child\,d)]\\ÊÊ{\Rightarrow} [|[p | (\pizz\,p) \land (((\eat\,e)p) d) {\land} ((\patient\,e) p) ] {\land} ((\agent\,e) d)]]$ 

%
%
%
%
%

If we add the $[e| \past(e)\land \cdots$ that has been left out, 
we obtain: $\exists e\ \past (e) \land \forall d \child(d) \Rightarrow \agent(e,d)\land \exists p (\pizz(p) \land \eat(e,p,d) \land \patient(e,p)) $ 

%

\section{Perspectives} The purpose of this paper was to describe a state of our work on categorial representation of generative grammars, which enables the automated computation of semantic representation, here viewed as DRSs. Nevertheless, some intriguing questions remain. 

On the syntactic side, what would be a proof theoretical version of shortest move? As such conditions involve the history of proofs, that are proof terms, 
dependant types should provide a way to formulate such condition, but for linear logic they have not yet been much studied. 
Still on the syntactic side, representation of minimalist grammars with remnant movement (which leaves out head-movement) for which the correspondence is tighter  by our work should be developed on particular phenomena, like VP rolls from Hungarian: as derivation intends to be quite lengthy, possibly some predetermined sequences or rules should be used, if they are proof-theoretically meaningful. 

On  the semantic side, some aspects deserve further study and improvement. The internalisation process,  although harmless because it is performed just after parsing, is not to be recommended. Indeed, it depends on the name of the variable and  modifies variable binding. Correlated to this issue, rules that depend on the name of a bound variable, the one associated with a movable constituent, should be avoided. 

We nevertheless hope that this movement inside the categorial community will make colleagues happy, and especially Jim Lambek.

\bibliography{bigbiblio}

\end{document}